%% file: top.tex
\documentclass[10pt,twocolumn,letterpaper]{article}

\usepackage{3dv}
\usepackage{times}
\usepackage{epsfig}
\usepackage{graphicx}
\usepackage{amsmath}
\usepackage{amssymb}
\usepackage{xcolor}
\usepackage{caption}
\usepackage{subcaption}
\usepackage{algorithm}
\usepackage[noend]{algpseudocode}
\usepackage[pagebackref=true,breaklinks=true,letterpaper=true,colorlinks,bookmarks=false]{hyperref}

\usepackage{array}
\usepackage{booktabs}
\usepackage{multirow}
\newcolumntype{L}[1]{>{\raggedright\let\newline\\\arraybackslash\hspace{0pt}}m{#1}}
\newcolumntype{C}[1]{>{\centering\let\newline\\\arraybackslash\hspace{0pt}}m{#1}}
\newcolumntype{R}[1]{>{\raggedleft\let\newline\\\arraybackslash\hspace{0pt}}m{#1}}

\graphicspath{ {gfx/} }

\input{commands}

\input{oc_math}
\threedvfinalcopy %

\def\threedvPaperID{97} %

\ifthreedvfinal\fi
\begin{document}

\title{OctNetFusion: Learning Depth Fusion from Data}

\author{Gernot Riegler$^1$ \qquad Ali Osman Ulusoy$^2$ \qquad Horst Bischof$^1$ \qquad Andreas Geiger$^{2,3}$\\
$^1$Institute for Computer Graphics and Vision, Graz University of Technology\\%
$^2$Autonomous Vision Group, MPI for Intelligent Systems T\"ubingen\\%
$^3$Computer Vision and Geometry Group, ETH Zürich\\%
{\tt\small \{riegler, bischof\}@icg.tugraz.at} \quad {\tt\small \{osman.ulusoy,andreas.geiger\}@tue.mpg.de}
}

\maketitle

\setlength\arraycolsep{1.5pt}

\input{sec_abstract.tex}

\input{sec_introduction.tex}

\input{sec_related.tex}

\input{sec_method.tex}

\input{sec_evaluation.tex}
\input{sec_conclusion.tex}

{\small
\bibliographystyle{ieee}
\bibliography{bibliography_long,bibliography}
}

\end{document}

%% file: commands.tex
\newcommand{\figref}[1]{\Fig~\ref{#1}}
\newcommand{\secref}[1]{Section~\ref{#1}}

\newcommand{\tabref}[1]{Table~\ref{#1}}

\makeatletter
\newcommand{\raisemath}[1]{\mathpalette{\raisem@th{#1}}}
\newcommand{\raisem@th}[3]{\raisebox{#1}{$#2#3$}}
\makeatother

\makeatletter
\DeclareRobustCommand\onedot{\futurelet\@let@token\@onedot}
\def\@onedot{\ifx\@let@token.\else.\null\fi\xspace}
\def\eg{e.g\onedot} 
\def\ie{i.e\onedot}

\def\wrt{wrt\onedot}

\def\etal{et~al\onedot} 
\def\Fig{Fig\onedot}   
\makeatother

\newcommand{\boldparagraph}[1]{\vspace{0.2cm}\noindent{\bf #1:} }

\definecolor{darkgreen}{rgb}{0,0.7,0}

%% file: sec_abstract.tex
\begin{abstract}
In this paper, we present a learning based approach to depth fusion, i.e., dense 3D reconstruction from multiple depth images. 
The most common approach to depth fusion is based on averaging truncated signed distance functions, which was originally proposed by Curless and Levoy in 1996.
While this method is simple and provides great results, it is not able to reconstruct (partially) occluded surfaces and requires a large number frames to filter out sensor noise and outliers. 
Motivated by the availability of large 3D model repositories and recent advances in deep learning, we present a novel 3D CNN architecture that learns to predict an implicit surface representation from the input depth maps.
Our learning based method significantly outperforms the traditional volumetric fusion approach in terms of noise reduction and outlier suppression.
By learning the structure of real world 3D objects and scenes, our approach is further able to reconstruct occluded regions and to fill in gaps in the reconstruction.
We demonstrate that our learning based approach outperforms both vanilla TSDF fusion as well as TV-L1 fusion on the task of volumetric fusion.
Further, we demonstrate state-of-the-art 3D shape completion results.
\end{abstract}

%% file: sec_introduction.tex
\section{Introduction} 
\label{sec:introduction}

Reconstructing accurate and complete 3D surface geometry is a core problem in computer vision. While image based techniques \cite{Seitz2006CVPR,Furukawa2010PAMI,Delaunoy2014CVPR,Ulusoy2015THREEDV,Schoenberger2016CVPR} provide compelling results for sufficiently textured surfaces, the introduction of affordable depth sensors, in particular the Microsoft Kinect sensor, allows scanning a wide variety of objects and scenes. This has led to the creation of large databases of real-world 3D content \cite{Choi2016ARXIV,Dai2017CVPR,Silberman2012ECCV,Song2015CVPR}, enabling progress in a variety of areas including 3D modeling \cite{Maturana2015IROS,Wu2015CVPR}, 3D reconstruction \cite{Choi2015CVPR,Whelan2015RSS}, 3D recognition \cite{Gupta2015CVPR,Song2014ECCV} and 3D scene understanding \cite{Geiger2015GCPR,Silberman2012ECCV}.

\begin{figure}[t!]
\begin{subfigure}{0.47\linewidth}
\includegraphics[width=\linewidth]{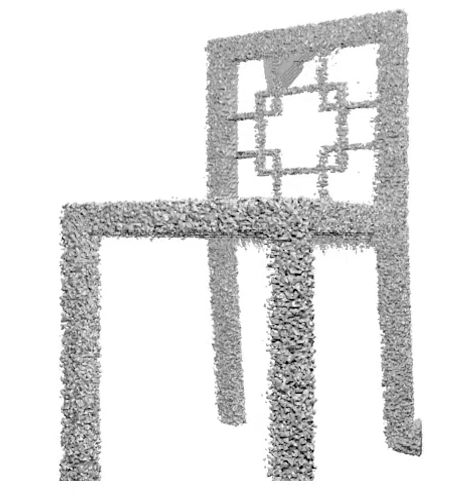}
\caption{TSDF Fusion~\cite{Curless1996SIGGRAPH}}
\end{subfigure}
\begin{subfigure}{0.47\linewidth}
\includegraphics[width=\linewidth]{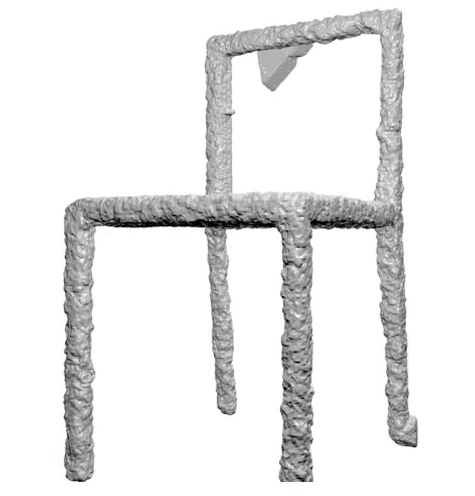}
\caption{TV-L1 Fusion~\cite{Zach2007ICCV}}
\end{subfigure}
\begin{subfigure}{0.47\linewidth}
\includegraphics[width=\linewidth]{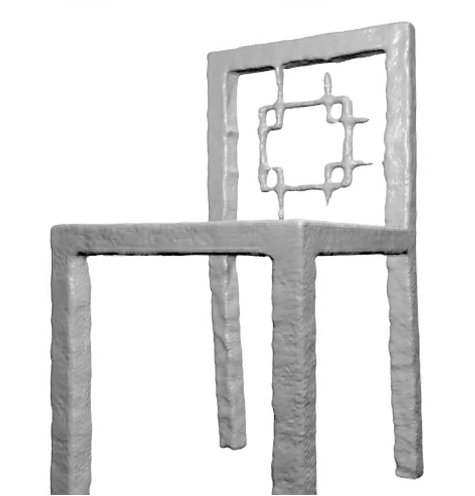}
\caption{Our Approach}
\end{subfigure}
\begin{subfigure}{0.47\linewidth}
\includegraphics[width=\linewidth]{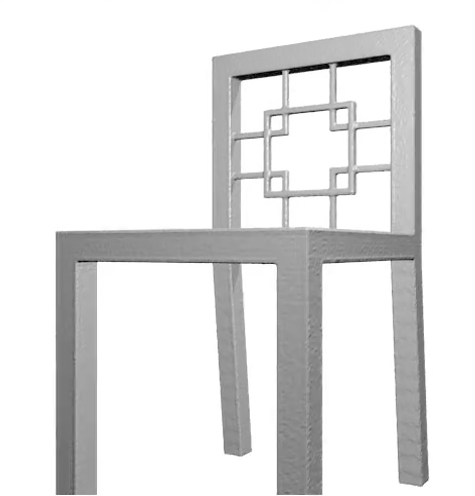}
\caption{Ground Truth}
\end{subfigure}
\caption{
{\bf Depth fusion results} using four uniformly spaced views around the object. (\textcolor{red}{a}) TSDF fusion produces noisy results. (\textcolor{red}{b}) TV-L1 fusion smoothes the shape, but also removes thin details. (\textcolor{red}{c}) Our approach learns from large 3D repositories and significantly improves fusion results in terms of noise reduction and surface completion.}
\label{fig:illustration}
\vspace{-1.2em}
\end{figure}

However, creating complete and accurate 3D models from 2.5D depth images remains a difficult problem. Challenges include sensor noise, quantization artifacts, outliers (\eg, bleeding at object boundaries) and missing data such as occluded surfaces. Some of these challenges can be addressed by integrating depth information from multiple viewpoints in a volumetric representation. In particular, Curless and Levoy demonstrate that averaging truncated signed distance functions (TSDF) allows for a simple yet effective approach to depth fusion~\cite{Curless1996SIGGRAPH} which is used in a large number of reconstruction pipelines today \cite{Choi2015CVPR,Newcombe2011ISMAR,Niesner2013SIGGRAPH,Steinbrucker2013ICCV,Whelan2012RSSWORK}.

Despite its popularity, TSDF fusion has two major drawbacks: First, it requires a large number of frames to smooth out sensor noise and outliers. Second, it does not allow for reconstructing occluded regions or completing large holes. 

In this paper, we tackle these problems by proposing a learning-based solution to volumetric 3D fusion. By utilizing large datasets of 3D models and high capacity 3D convolutional neural networks (3D CNNs), our approach learns to smooth out sensor noise, deals with outliers and completes missing 3D geometry. 
To the best of our knowledge, our approach is the first to learn volumetric fusion from noisy depth observations. 

3D CNNs \cite{Choy2016ECCV,Huang2016ICPR,Maturana2015IROS,Qi2016CVPR,Wu2016ECCV} are a natural choice for formulating this task as an end-to-end learning problem. However, existing deep learning approaches are limited to small resolutions (typically $32^3$ voxel grids) due to the cubic growth in memory requirements. A notable exception is the OctNet approach of Riegler \etal \cite{Riegler2017CVPR} which takes advantage of the sparsity of 3D volumetric models using an octree representation \cite{Samet2005}, enabling deep learning at resolutions of $256^3$ voxels and beyond.

The main limitation of OctNets \cite{Riegler2017CVPR}, however, is that the octree representation is derived from the {\it input} and fixed during learning and inference. While this is sufficient for tasks such as 3D classification or semantic segmentation where the input and the output share the same octree representation, the OctNet framework does not directly apply to tasks where the 3D space partitioning of the output is {\it unknown} a priori and may be different than that of the input. In particular, for tasks such as depth map fusion and 3D completion the location of the implicit surface is unknown and needs to be inferred from noisy observations.

The key contribution of this paper is to lift this restriction.
More specifically, we propose a novel 3D CNN architecture termed {\it OctNetFusion} which takes as input one or more depth images and estimates both the 3D reconstruction and its supporting 3D space partitioning, \ie the octree structure of the {\it output}.
We apply this architecture to the depth map fusion problem and formulate the task as the prediction of truncated signed distance fields which can be meshed using standard techniques~\cite{Lorensen1987SIGGRAPH}.

We evaluate our approach on synthetic and real-world datasets, studying several different input and output representations, including TSDF.
Our experiments demonstrate that the proposed method is able to reduce noise and outliers compared to vanilla TSDF fusion~\cite{Curless1996SIGGRAPH,Newcombe2011ISMAR} while avoiding the shrinking bias of local regularizers such as TV-L1 \cite{Zach2007ICCV}.
Besides, our model learns to complete missing surfaces and fills in holes in the reconstruction.
We demonstrate the flexibility of our model by evaluating it on the task of volumetric shape completion from a single view where we obtain improvements \wrt the state-of-the-art \cite{Firman2016CVPR}. 
Our code is on GitHub: \url{https://github.com/griegler/octnetfusion}.

%% file: sec_related.tex
\section{Related Work} 
\label{sec:related}

\boldparagraph{Volumetric Fusion}
In their seminal work, Curless and Levoy \cite{Curless1996SIGGRAPH} proposed to integrate range information across viewpoints by averaging truncated signed distance functions. The simplicity of this method has turned it into a universal approach that is used in many 3D reconstruction pipelines. Using the Microsoft Kinect sensor and GPGPU processing, Newcombe \etal \cite{Newcombe2011ISMAR} showed that real-time 3D modeling is feasible using this approach.
Large-scale 3D reconstruction has been achieved using iterative re-meshing \cite{Whelan2012RSSWORK} and efficient data structures \cite{Niesner2013SIGGRAPH,Steinbrucker2013ICCV}. The problem of calibration and loop-closure detection has been considered in \cite{Choi2015CVPR,Zhou2013ICCV}.
Due to the simplicity of the averaging approach, however, these methods typically 
require a large number of input views, are susceptible to outliers in the input and don't allow to predict surfaces in unobserved regions.

Noise reduction can be achieved using variational techniques which integrate local smoothness assumptions \cite{Blaha2016CVPR,Haene2013CVPR,Zach2007ICCV} into the formulation. However, those methods are typically slow and can not handle missing data. In this paper, we propose an alternative learning based solution which significantly outperforms vanilla TSDF fusion \cite{Curless1996SIGGRAPH,Newcombe2011ISMAR} and TV-L1 fusion \cite{Zach2007ICCV} in terms of reconstruction accuracy.

\boldparagraph{Ray Consistency}
While TSDF fusion does not explicitly consider free space and visibility constraints, ray potentials allow for modeling these constraints in a Markov random field.  Ulusoy \etal \cite{Ulusoy2015THREEDV,Ulusoy2017CVPR} consider a fully probabilistic model for image based 3D reconstruction. Liu and Cooper \cite{Liu2014PAMI} formulate the task as MAP inference in a MRF. 
In contrast to our method, these algorithms do not learn the geometric structure of objects and scene from data. Instead, they rely on simple hand-crafted priors such as spatial smoothness~\cite{Liu2014PAMI}, or piecewise planarity~\cite{Ulusoy2016CVPR}. Notably, Savinov \etal combine ray potentials with 3D shape regularizers that are learned from data~\cite{Savinov2015CVPR}. However, their regularizer is {\it local} and relies on a semantic segmentation as input. In this work, we do not consider the semantic class of the reconstructed object or scene and focus on the generic 3D reconstruction problem using a {\it global} model.

\boldparagraph{Shape Completion}
If exact 3D models are available, missing surfaces can be completed by detecting the objects and fitting 3D models to the observations \cite{Brachmann2014ECCV,Hinterstoisser2012ACCV}. In this paper, we assume that such prior knowledge is not available. Instead we directly learn to predict the 3D structure from training data in an end-to-end fashion.

Shape completion from a single RGB-D image has been tackled in \cite{Firman2016CVPR,Kim2013ICCV,Song2017CVPR,Zheng2013CVPR}. While \cite{Kim2013ICCV} use a CRF for inference, \cite{Firman2016CVPR} predict structured outputs using a random forest and \cite{Song2017CVPR} use a CNN to jointly estimate voxel occupancy and semantic class labels.
In contrast to our approach, these methods reason at the voxel level and therefore do not provide sub-voxel surface estimates.
Furthermore, existing 3D CNNs \cite{Song2017CVPR} are limited in terms of resolution.
In this paper, we demonstrate a unified approach which allows to reason about missing 3D structures at large resolution while providing sub-voxel surface estimates. In contrast to single-image reconstruction methods, our approach naturally handles an arbitrary number of input views. For the task of 3D shape completion from a single image we obtain results which are superior to those reported by Firman \etal \cite{Firman2016CVPR}.

In very recent work, Dai \etal \cite{Dai2017CVPRa} consider the problem of high-resolution 3D shape completion.
Their approach first regresses $32^3$ voxel volumes using a 3D CNN, followed by a multi-resolution 3D shape synthesis step using a large database of 3D CAD models \cite{Chang2015ARXIV}.
While their object-centric approach is able to reconstruct details of individual objects with known 3D shape, we put our focus on general 3D scenes where such knowledge is not available.

%% file: sec_method.tex
\begin{figure*}[t]
\center
\begin{subfigure}{\linewidth}
  \centering
  \includegraphics[width=0.95\linewidth]{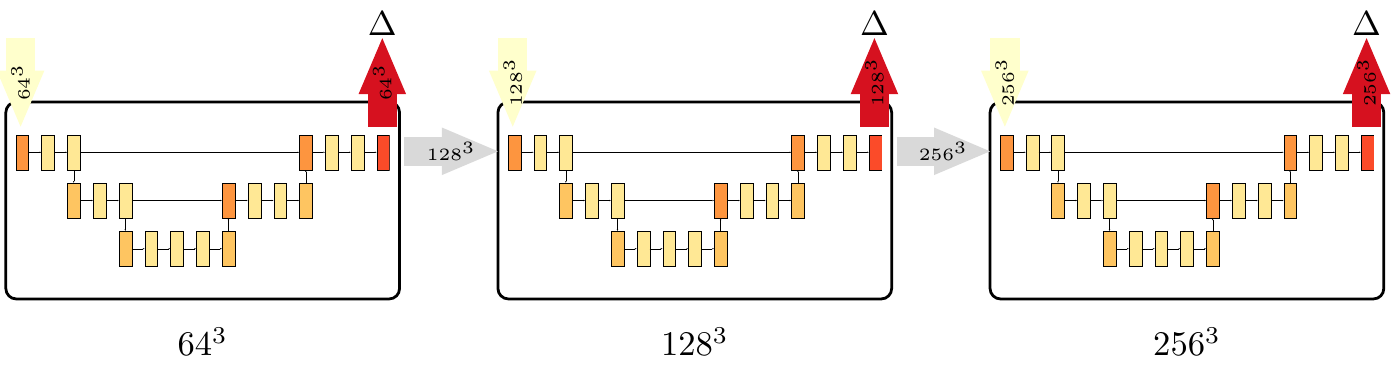}
  \caption{\textbf{OctNetFusion Architecture}}
  \label{fig:octnet_fusion_network}
\end{subfigure}
\begin{subfigure}{0.9\linewidth}
  \includegraphics[width=\linewidth]{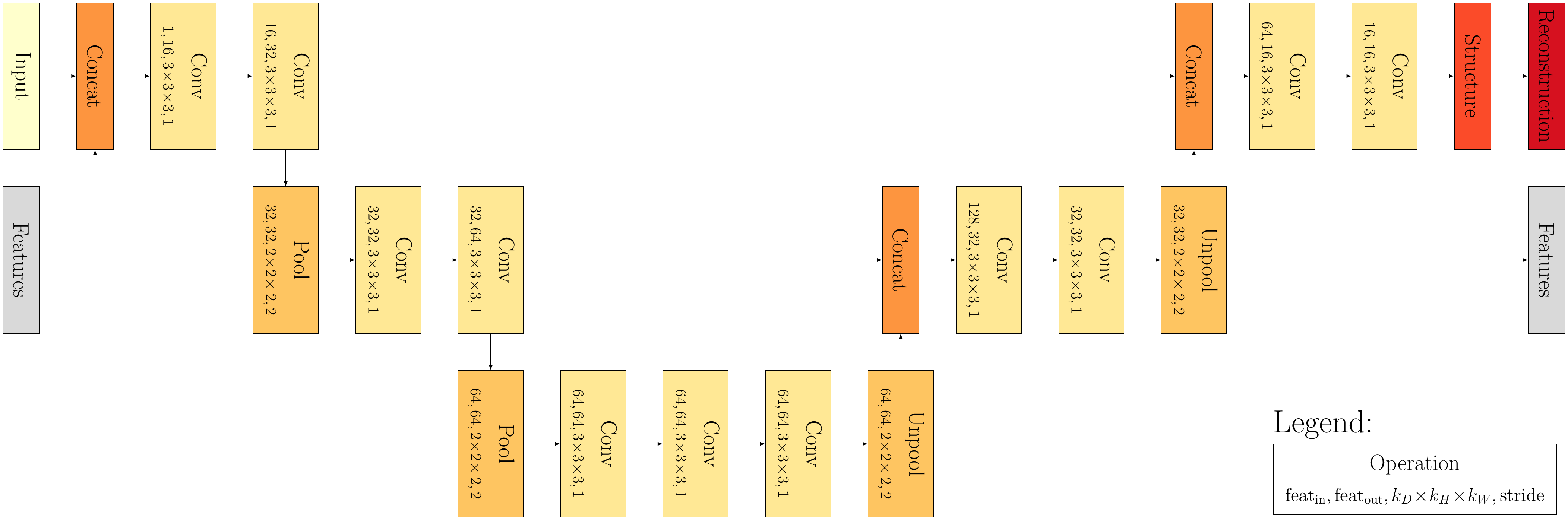}
  \caption{\textbf{Encoder-Decoder Module}}
  \label{fig:octnet_fusion_encoder_decoder_module}
\end{subfigure}
\caption{\textbf{OctNetFusion Architecture.} (\subref{fig:octnet_fusion_network}) Overall structure of our coarse-to-fine network.
 (\subref{fig:octnet_fusion_encoder_decoder_module}) Each encoder-decoder module increases the receptive field size and adds contextual information. The structure module (orange) is detailed in \figref{fig:octnet_fusion_structure_module}.}
  \label{fig:octnet_fusion}
\end{figure*}

\begin{figure}[t!]
	\center
	\includegraphics[width=\linewidth]{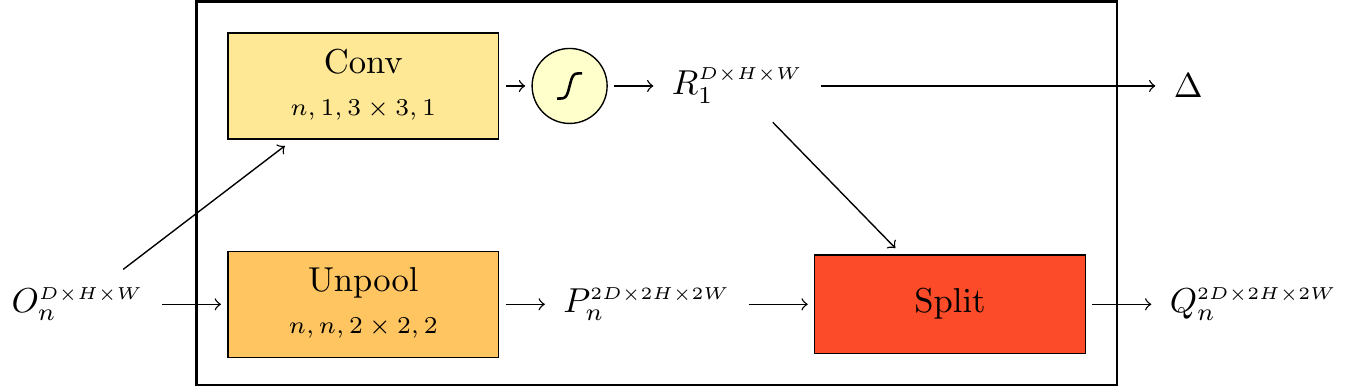}
	\caption{\textbf{Structure Module}. The structure manipulation module doubles the spatial resolution of the feature maps. A loss at the end of each pyramid level ($\Delta$) measures the quality of the reconstruction at the respective resolution.} %
	\label{fig:octnet_fusion_structure_module}
\end{figure}

\section{Method} 
\label{sec:method}

This section introduces our OctNetFusion architecture.
Our work builds upon the recent work of 3D octree convolutional networks~\cite{Riegler2017CVPR,Wang2017SIGGRAPH}.
As our work specifically extends~\cite{Riegler2017CVPR}, we follow its notation whenever possible. 
To make this paper self-contained, we first briefly review OctNet~\cite{Riegler2017CVPR} in \secref{sec:octnet}.
Then, we present our OctNetFusion approach in \secref{sec:octnetfusion} which learns to jointly estimate the output quantity (\eg, signed distance or occupancy) and the space partitioning.
\secref{sec:input_output_encoding} specifies the feature representations considered in our experiments.

\subsection{OctNet}
\label{sec:octnet}

The main limitation of conventional 3D CNNs that operate on regular voxel grids is the cubic growth in memory requirements with respect to the voxel resolution. 
However, 3D data is often sparse in nature \cite{Li2016ARXIV}. 
For instance, the surface of an object can be interpreted as a 2D manifold in 3D space.
Riegler \etal \cite{Riegler2017CVPR} utilize this observation and define a CNN on the grid-octree data structure of \cite{Miller2011GPGPU}.
The data structure itself consists of a grid of shallow octrees with maximum depth $D=3$, trading off computation and memory.
The structure of the shallow octrees can be efficiently encoded as bit strings that allows for rapid retrieval of neighboring cells.
One important property of OctNets is that none of the operations (\ie, convolution, pooling, unpooling) changes the grid-octree data structure which is based on the input (\eg, point cloud, voxel grid).
This can be seen as a data-adaptive pooling operation which maps the output of each layer back to the grid-octree representation.

We now introduce the basic notation. Consider a dense voxel grid $T \in \mathbb{R}^{8D \times 8H \times 8W}$ where $T_{i,j,k}$ denotes the value at voxel $(i,j,k)$.
Further, let $O$ denote a grid-octree data structure that covers the same volume.
Given tree depth of $D=3$ this data structure would contain $D \times H \times W$ shallow octrees, where each shallow octree covers $8^3$ voxels. 
The important difference to the dense representation is that the cells in $O$ can comprise a variable number of voxels. 

Let $\Omega[i,j,k]$ denote the smallest grid-octree cell that contains the voxel at $(i,j,k)$. 
$\Omega[i,j,k]$ can be interpreted as the set of voxel indices, whose data is pooled to a single value as described above.
Furthermore, $|\Omega[i,j,k]|$ denotes the number of voxels comprised by this cell.
If the cell is at the finest resolution of the tree, we have $|\Omega[i,j,k]| = 1$, \ie, the cell is equal to the voxel in $T_{i,j,k}$.
In contrast, if the complete shallow octree consists of only a single leaf cell, then $|\Omega[i,j,k]| = 512$ as all $8^3$ voxels are pooled.

Given this basic notation, the authors of \cite{Riegler2017CVPR} show how the convolution, pooling and unpooling operation can be efficiently implemented on this data structure. We refer to \cite{Riegler2017CVPR} for further details.

\subsection{OctNetFusion}
\label{sec:octnetfusion}

The main drawback of OctNets \cite{Riegler2017CVPR} is that the octree structure of the input and output, \ie the partitioning of the 3D space, has to be known a priori. 
This is a reasonable assumption for tasks like 3D point cloud labeling (\eg, semantic segmentation) where the input and the output octree structures are the same.
However, for tasks where the output geometry is different from the input geometry, \eg, in volumetric fusion or shape completion, the grid-octree data structure has to be adapted during inference. 

We now present our OctNetFusion architecture, illustrated in \figref{fig:octnet_fusion}, which allows to learn the grid-octree structure along with the 3D task in a principled manner.

\boldparagraph{Network Architecture}
Our overall network architecture is illustrated in \figref{fig:octnet_fusion_network}. We represent the voxelized input and output using the grid-octree structure described in \secref{sec:octnet}. The input to the network is a feature volume (\eg, TSDF), calculated from a single or multiple depth maps, see \secref{sec:input_output_encoding} for details. The output may encode a TSDF or a binary occupancy map, depending on the application.

As the 3D input to our method can be sparse and incomplete, we refrain from using the classical U-shaped architecture as common for 2D-to-2D prediction tasks \cite{Badrinarayanan2015ARXIV,Dosovitskiy2015ICCV}. Instead, we propose a coarse-to-fine network with encoder-decoder modules, structure manipulation modules and a loss defined at every pyramid level. More specifically, we create a 3D scale pyramid where the number of voxels along each dimension increases by a factor of two between pyramid levels. At each level, we process the input using an encoder-decoder module which enlarges the receptive field and captures contextual information. We pass the resulting features to a structure manipulation module which computes the output at the respective resolution, increases the resolution and updates the structure of the network for further processing.
We propagate features to successively finer resolutions until we have reached the final target resolution. %
We will now describe the encoder-decoder module and the structure module in detail.

\boldparagraph{Encoder-Decoder Module}
The encoder-decoder module is illustrated in \figref{fig:octnet_fusion_encoder_decoder_module}. It combines convolution layers with pooling and unpooling layers similar to the segmentation network used in \cite{Riegler2017CVPR}.
All convolutional layers are followed by a ReLU non-linearity \cite{Nair2010ICML}.
The convolution layer before each pooling operation doubles the number of feature maps while the convolution layer after each unpooling operation halves the number of features.
Pooling operations reduce spatial information but increase the level of context captured in the features.
The result of the unpooling operation is concatenated with the corresponding high-resolution features from the encoder path to combine high-resolution information with low-resolution contextual cues.

\boldparagraph{Structure Module}
As discussed above, the unpooling operation of the original OctNet architecture \cite{Riegler2017CVPR} has one major drawback: the octree structure must be known in advance to determine which voxels shall be split. While for 3D point labeling the structure can be split according to the input, the final output structure is unknown for tasks like volumetric fusion or completion.
Na\"{i}vely splitting all voxels would eliminate the advantage of the data-adaptive representation and limit the output resolution to small volumes.

Consequently, we introduce a {\it structure module} after each encoder-decoder module which determines for each voxel if it shall be split (\ie, close to the surface) or not (\ie, far from the surface).
Our structure module is illustrated in \figref{fig:octnet_fusion_structure_module}.
The main idea is to add a split mask to the standard unpooling operation that indicates which of the octree cells should be further subdivided. 
This splitting mask is then used to subdivide the unpooled grid-octree structure.

More formally, let us consider an input grid-octree structure $O$ with $n$ feature channels and $D \times H \times W$ shallow octrees as illustrated in \figref{fig:octnet_fusion_structure_module}.
After the unpooling operation we obtain a structure $P$ that consists of  $2D \times 2H \times 2W$ shallow octrees where each octree cell comprises eight-times the number of voxels, \ie, $|\Omega_P[2i,2j,2k]| = 8 |\Omega_O[i,j,k]|$.

To determine the new octree structure, we additionally predict a reconstruction $R$ at the resolution of $O$ using a single convolution followed by a sigmoid non-linearity or a $1\times 1$ convolution depending on the desired output (occupancy or TSDF, respectively).
A reconstruction loss ($\Delta$) ensures that the predicted reconstruction is close to the ground truth reconstruction at each resolution of the scale pyramid (for learning the network we provide the ground truth reconstruction at each resolution as described in \secref{sec:evaluation}).

We define the split mask implicitly by the surface of the reconstruction $R$.
The surface is defined by the gradients of $R$ when predicting occupancies or by the zero-levelset of $R$ in the case of TSDF regression.
Given the surface, we split all voxels within distance $\tau$ from the surface.
For TSDF regression, $\tau$ equals the truncation threshold.
For occupancy classification, $\tau$ is a flexible parameter which can be tuned to trade reconstruction accuracy vs. memory usage.
The output of this split operation finally yields the high resolution structure $Q$ which serves as input to the next level.

\subsection{Input / Output Encoding}
\label{sec:input_output_encoding}

This section describes the input and output encodings for our method. 
An ablation study, analyzing the individual encodings is provided in \secref{sec:evaluation}.

\subsubsection{Input Encoding}
\label{sec:input_encoding}
The input to our method are one or more 2.5D depth maps. %
We now discuss several ways to project this information into 3D voxel space which, represented using grid-octree structures, forms the input to the OctNetFusion architecture described above.
The traditional volumetric fusion approach \cite{Curless1996SIGGRAPH} calculates the weighted average TSDF with respect to all depth maps independently for every voxel where the distance to the surface is measured along the line of sight to the sensor.
While providing for a simple one-dimensional signal at each voxel, this encoding does not capture all information due to the averaging operation.
Thus, we also explore higher dimensional input encodings which might better retain the information present in the sensor recordings.
We now formalize all input encodings used during our experiments, starting with the simplest one.

\boldparagraph{Occupancy Fusion (1D)}
The first, and simplest encoding fuses information at the occupancy level.
Let $d_v^{(i)}$ be the depth of the center of voxel $v$ \wrt camera $i$.
Further, let $d_c^{(i)}$ be the value of the depth map when projecting voxel $v$ onto the image plane of camera $i$.
Denoting the signed distance between the two depth values $\delta_v^{(i)} = d_c^{(i)}-d_v^{(i)}$, we define the occupancy of each voxel $o(v)$ as
\begin{align}
  o(v) = \begin{cases}
    1 & \exists i: \delta_v^{(i)} \le s \land \nexists i: \delta_v^{(i)} > s \\
    0 & \text{else}
  \end{cases}
\end{align}
where $s$ is the size of voxel $v$.
The interpretation is as follows:
If there exists any depth map in which voxel $v$ is observed as free space the voxel is marked as free, otherwise it is marked as occupied. 
While simple, this input encoding is susceptible to outliers in the depth maps and doesn't encode uncertainty. 
Furthermore, the input distance values are not preserved as only occupancy information is encoded.

\boldparagraph{TSDF Fusion (1D)}
Our second input encoding is the result of traditional TSDF fusion as described in \cite{Curless1996SIGGRAPH,Newcombe2011ISMAR}. 
More specifically, we project the center of each voxel into every depth map, calculate the TSD value using a truncation threshold $\tau$ (corresponding to the size of four voxels in all our experiments), and average the result over all input views. 
While various weight profiles have been proposed \cite{Steinbrucker2013ICCV}, we found that the simple constant profile proposed by Newcombe \etal \cite{Newcombe2011ISMAR} performs well.
This input representation is simple and preserves distances, but it does not encode uncertainty and thus makes it harder to resolve conflicts.

\boldparagraph{TDF + Occupancy Fusion (2D)}
The TSDF encoding can also be split into two channels:
one channel that encodes the truncated unsigned distance to the surface (TDF) and one that encodes occupancy.
Note that, if $t(v)$ is the TDF of voxel $v$, and $o(v)$ its occupancy, then $-t(v) \cdot o(v)$ is equivalent to the truncated signed distance function of $v$.

\boldparagraph{Histogram (10D)}
While the previous two encodings captures surface distance and uncertainty, they do not maintain the multi-modal nature of fused depth measurements. 
To capture this information, we propose a histogram-based representation. 
In particular, we encode all distance values for each voxel using a 10D histogram with 5 bins for negative and 5 bins for positive distance values. 
The first and the last bin of the histogram capture distance values beyond the truncation limit, while the bins in between collect non-truncated distance values. 
To allow sub-voxel surface estimation, we choose the histogram size such that a minimum of 2 bins are allocated per voxel. 
Furthermore, we populate the histogram in a smooth fashion by distributing the vote of each observation linearly between the two closest bins, \eg, we assign half of the mass to both neighboring bins if the prediction is located at their boundary.

\subsubsection{Output Encoding and Loss}

Finally, we describe the output encodings and the loss we use for the volumetric fusion and the volumetric completion tasks we consider in the experimental evaluation.

\boldparagraph{Volumetric Fusion}
For volumetric fusion, we choose the TSDF as output representation using an appropriate truncation threshold $\tau$. Note that in contrast to binary occupancy, TSDF outputs allow for predicting implicit surfaces with sub-voxel precision.
We regress the TSDF values at each resolution (\ie, within the structure module) using a $1\times 1$ convolution layer and use the $\ell_1$ loss for training.

\boldparagraph{Volumetric Completion}
For volumetric completion from a single view, we use a binary occupancy representation to match the setup of the baselines as closely as possible. Following common practice, we leverage the binary cross entropy loss for training the network.

%% file: sec_evaluation.tex
\section{Evaluation} 
\label{sec:evaluation}

In this section, we present our experiments and evaluations. 
In \secref{sec:volumetric_fusion} we consider the task of volumetric fusion from multiple depth images and in \secref{sec:volumetric_completion} we compare our approach to a state-of-the-art baseline on the task of volumetric completion from a single depth image.

\subsection{Volumetric Fusion}
\label{sec:volumetric_fusion}

In this section we consider the volumetric fusion task. We evaluate our OctNetFusion approach on the synthetic ModelNet40 dataset of Wu \etal \cite{Wu2015CVPR} as well as on real Kinect object scans that are generated using the depth image dataset by Choi \etal \cite{Choi2016ARXIV}.

Unfortunately, the ground truth TSDF can not be calculated directly from the 3D models in ModelNet as the meshes are not watertight, \ie, they contain holes and cracks.
Moreover, the meshes typically do not have consistently oriented normals.
Instead, we obtain the ground truth TSDF by densely sampling views around the object, rendering the input 3D model from all views and running traditional volumetric fusion on all generated (noise-free) depth maps. 
We found that $80$ views cover all object surfaces and hence allow for computing highly accurate TSDF ground truth.
For each of the categories we used $200$ models for training and $20$ for testing from the provided train/test split, respectively.

\begin{figure}
  \includegraphics[width=0.32\linewidth]{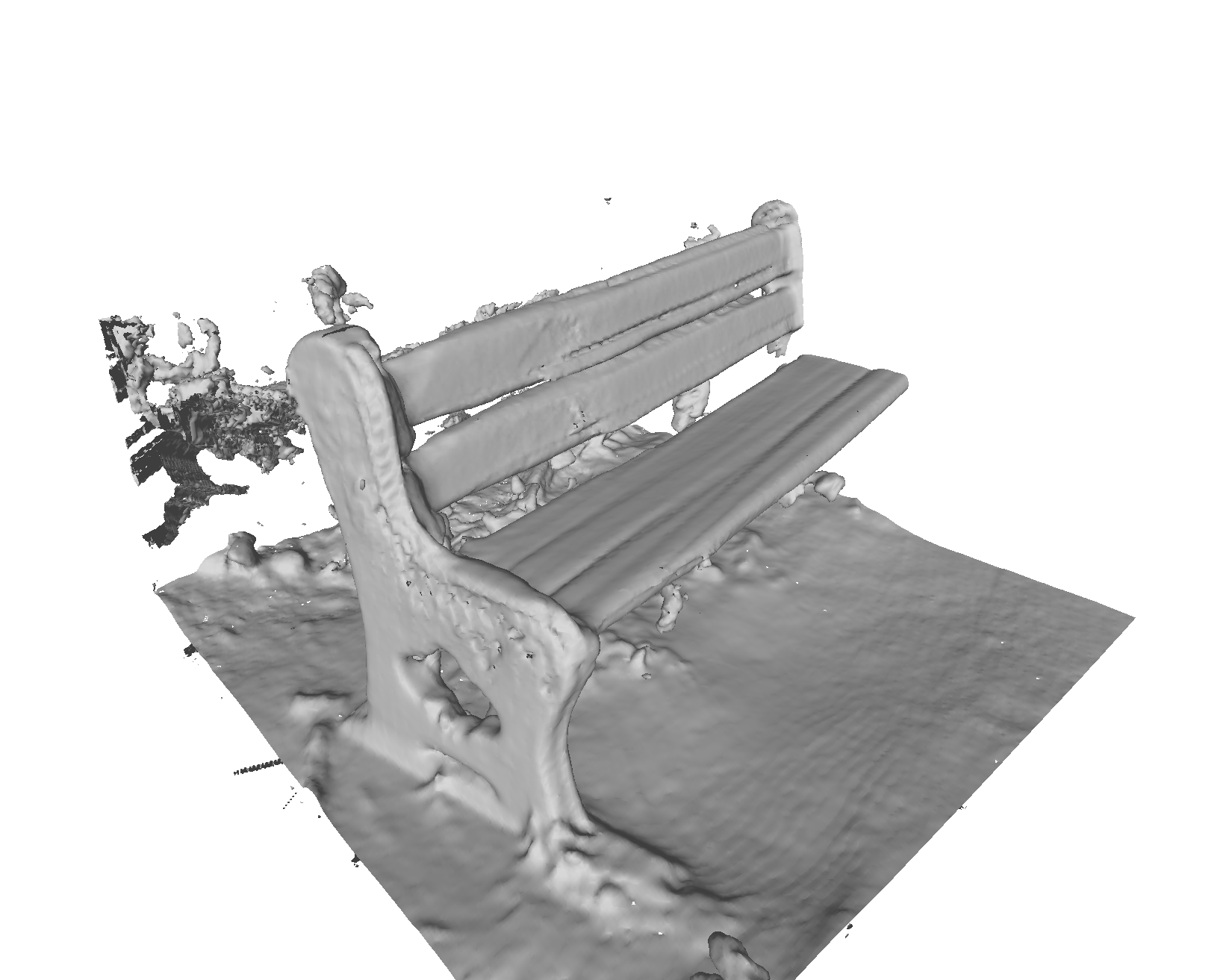}%
  \includegraphics[width=0.32\linewidth]{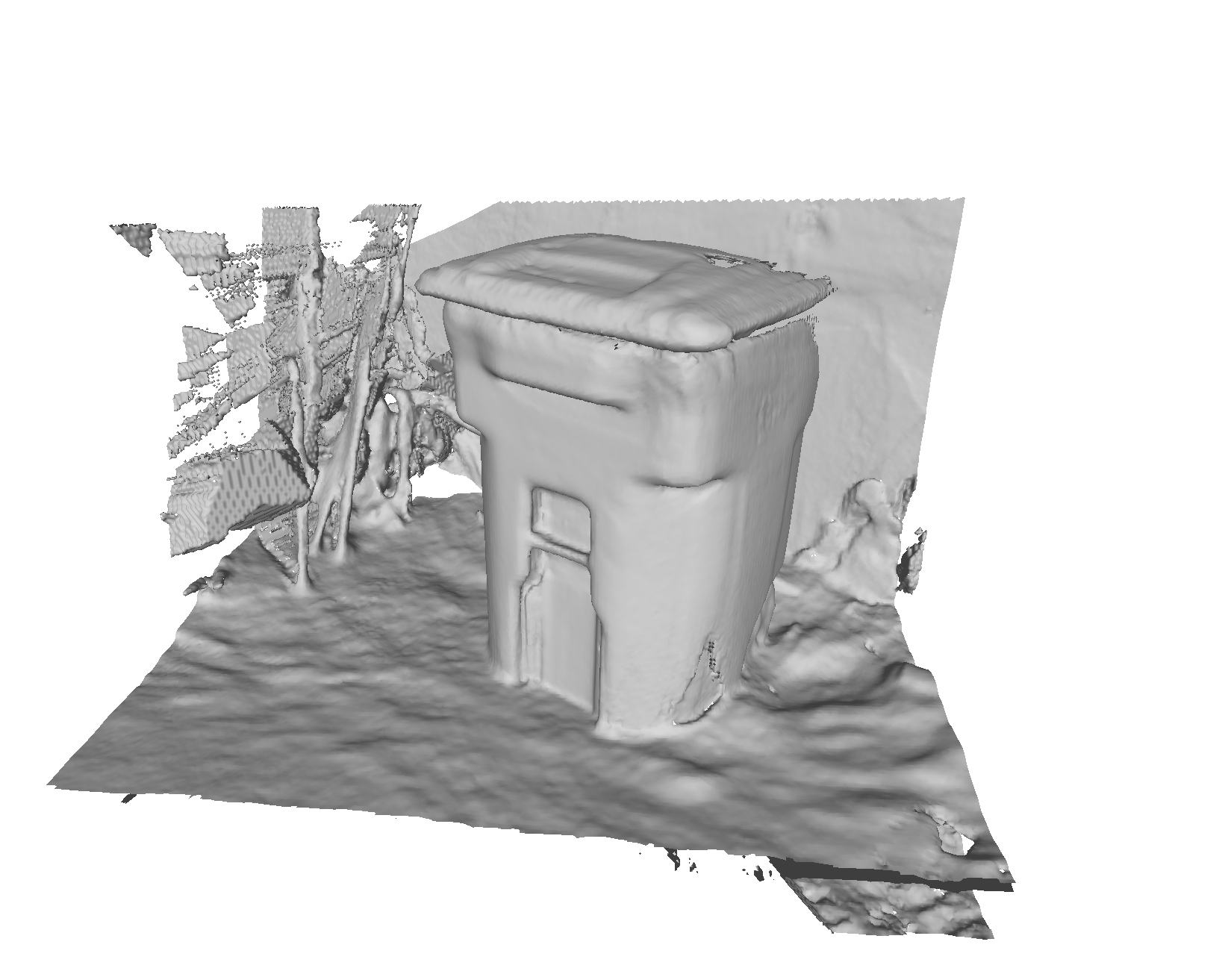}%
  \includegraphics[width=0.32\linewidth]{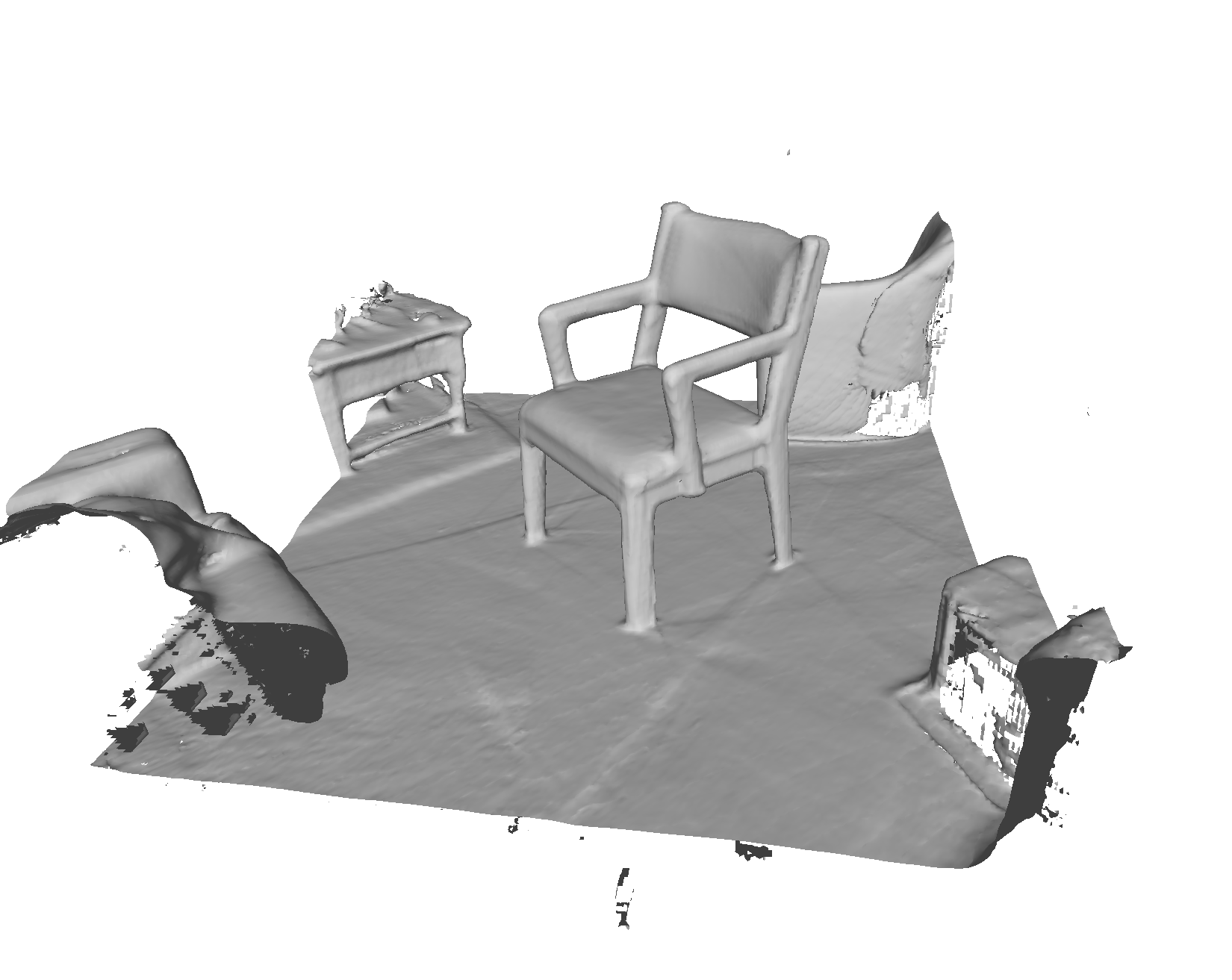}%
  \caption{{\bf Examples from Kinect Scan Dataset}}
  \label{fig:kinect_object_scans}
  \vspace{-1.2em}
\end{figure}

Besides the synthetic ModelNet dataset, we also evaluated our approach on real data from a Kinect RGB-D sensor. 
In particular, we use the $10$ videos captured by Choi \etal \cite{Choi2016ARXIV} which include a diverse set of objects such as chairs, tables, trash containers, plants, signs, etc. 
Unfortunately, the dataset does not include ground-truth 3D models or camera poses.
We thus estimated the camera poses using Kintinuous~\cite{Whelan2012RSSWORK} and visually inspect all models to remove those for which the pose tracking failed.
Similar to the ModelNet experiments, we leverage TSDF fusion to obtain reference 3D models. However, for this dataset we leverage $1000$ viewpoints for each object to average the effect of noise. This is possible as the dataset has been carefully collected with many slow camera motion and many redundant views. At test time we provide only a small fraction of views (10-20) to each algorithm to simulate challenging real-world conditions.
Example reference models produced by this procedure are shown in \figref{fig:kinect_object_scans}.
We augment the dataset by generating $20$ different view configurations per scene by selecting different and disjoint subsets of view points at random.

We train each stage of the network with a constant learning rate of $10^{-4}$ and Adam~\cite{Kingma2015ICLR} as optimizer.
We train the first stage for $50$ epochs, and the next two stages for $25$ epochs each.
We initialize the first stage according to the randomization scheme proposed in~\cite{He2015ICCV}, and initialize the weights of the other stages with those of the previous stage.

\begin{table}
  \center
    \center
    \setlength{\tabcolsep}{0.25em}
    {\footnotesize \input{tab/encodings/sad.tex} }
  \caption{{\bf Evaluation of Input Encodings} (MAD in mm)}
  \label{tab:mn_enc}
  \vspace{-0.8em}
\end{table}

\boldparagraph{Input Encoding}
We first investigate the impact of the input encodings discussed in \secref{sec:input_encoding} on the quality of the output. 
Towards this goal, we scaled the ModelNet objects to fit into a cube of $3\times 3 \times 3$ meters and rendered the depth maps onto $4$ equally spaced views sampled from a sphere. 
To simulate real data, we added depth dependent Gaussian noise to the inputs as proposed in~\cite{Park2011ICCV}. 

Our results are shown in \tabref{tab:mn_enc}. We compare the traditional volumetric fusion approach of Curless \etal~\cite{Curless1996SIGGRAPH} ("VolFus") and the variational approach of Zach \etal \cite{Zach2007ICCV} ("TV-L1") to our method using the input encodings described in \secref{sec:input_encoding}. The parameters of the baselines have been chosen based on cross-validation. We evaluate our results in terms mean absolute distance (MAD) which we compute over all voxels in the scene.
Each row shows results at a particular resolution, ranging from $64^3$ to $256^3$ voxels.

First, we observe that our model outperforms the traditional fusion approach \cite{Curless1996SIGGRAPH,Zach2007ICCV} as well as TV-L1 fusion \cite{Zach2007ICCV} by a large margin.
Improvements are particularly pronounced at high resolutions which demonstrates that our learning based approach is able to refine and complete geometric details which can not be recovered using existing techniques. 
Furthermore, we observe that the TSDF histogram based encoding yields the best results. 
We thus use this input encoding for all experiments that follow.

\begin{table}[t!]
  \centering
    {\footnotesize \input{tab/views/sad.tex} }
  \caption{{\bf Number of Input Views} (MAD in mm)}
  \label{tab:mn_views}
  \vspace{-1em}
\end{table}

\begin{table*}[ht]
	\center
	\setlength{\tabcolsep}{0.4em}
	{\footnotesize \input{tab/noise/sad.tex} }
	\caption{{\bf Evaluation wrt. Input Noise} (MAD in mm)}
	\label{tab:mn_noise}
  \vspace{-1em}
\end{table*}

\boldparagraph{Number of Views}
Next, we evaluate the performance of our network when varying the number of input views from one to six on the ModelNet dataset. All experiments are conducted at a resolution of $256^3$  voxels.
Our results are shown in \tabref{tab:mn_views}. 
Again, our approach outperforms both baselines in all categories. 
As expected, performance increases with the number of viewpoints. 
The largest difference between the baseline and our approach is visible for the experiment with only one input view. 
While no fusion is performed in this case, this demonstrates that our learned model is effective in completing missing geometry. 
When considering four input views, our approach reduces errors by a factor of $2$ to $3$ \wrt  TSDF fusion and TV-L1 fusion.

\boldparagraph{Noise on Input}
  In our next experiment we evaluate the impact of noise in the depth maps on the reconstruction.
  \tabref{tab:mn_noise} summarizes our results.
  We observe that our method is faithful \wrt the increase of input noise. 
  The MAD increases from $0.274$ mm for no noise to $0.374$ mm for severe noise ($\sigma=0.03$).
  In contrast, for TSDF fusion the MAD increases by more than $0.5$ mm and for TV-L1 fusion by more than $0.2$ mm.

\begin{table}
  \center
   {\footnotesize \input{tab/classgeneralization/sad.tex} }
  \caption{{\bf Seen vs. Unseen Categories} (MAD in mm)}
  \vspace{-1em}
  \label{tab:mn_class_aware}
\end{table}

\boldparagraph{Generalization on Unseen Categories}
Most existing approaches that leverage deep learning for 3D reconstruction train a model {\it specific} for a particular object class~\cite{Brock2016ARXIV,Choy2016ECCV,Girdhar2016ECCV,Sharma2016ARXIV,Rezende2016ARXIV} which typically does not generalize well to other classes or scenes with varying backgrounds as present in the scans of Choi \etal \cite{Choi2016ARXIV}. 
In contrast, here we are interested in 3D reconstruction of {\it general} scenes. 
We therefore analyze how our model behaves on shapes that where not seen during training.
In \tabref{tab:mn_class_aware} we trained a network on only $8$ categories out of $10$ and use the two unseen ones for testing ("Unseen"). We compare these results to the case where we train on all $10$ categories ("Seen"). Note that in neither case training shapes are part of the test set, but shapes from the same category are used or ignored. %
While we can observe a slight decrease in performance for the unseen categories, it is still far better than simple TSDF fusion, or TV-L1 fusion.

\begin{figure*}
\begin{subfigure}{0.5\linewidth}
	\setlength{\tabcolsep}{2pt}
  \begin{tabular}{@{} C{0.05\textwidth} C{0.2\textwidth} C{0.2\textwidth} C{0.2\textwidth} C{0.2\textwidth} C{0.01\textwidth}}
    \rotatebox{90}{\small{$64^3$}} & 
    \includegraphics[width=0.2\textwidth]{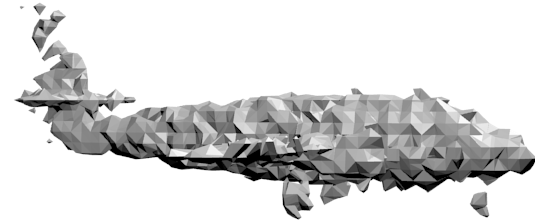} &
    \includegraphics[width=0.2\textwidth]{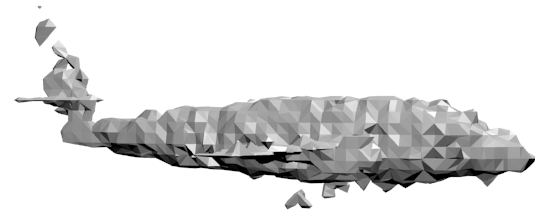} &
    \includegraphics[width=0.2\textwidth]{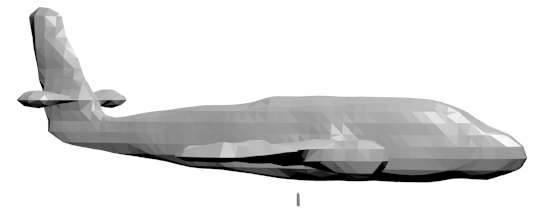} &
    \includegraphics[width=0.2\textwidth]{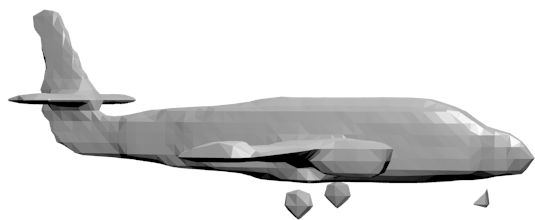} & \\[1.4cm]
    \rotatebox{90}{\small{$128^3$}} & 
    \includegraphics[width=0.2\textwidth]{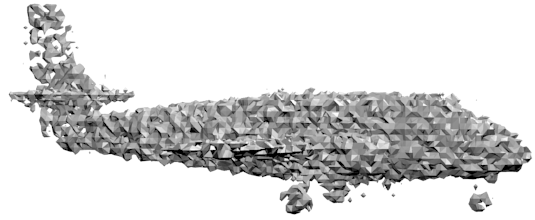} &
    \includegraphics[width=0.2\textwidth]{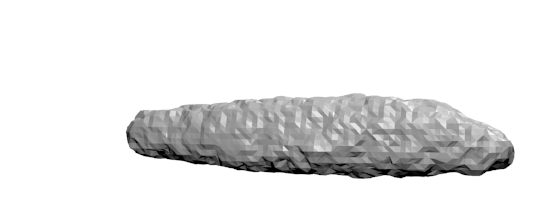} &
    \includegraphics[width=0.2\textwidth]{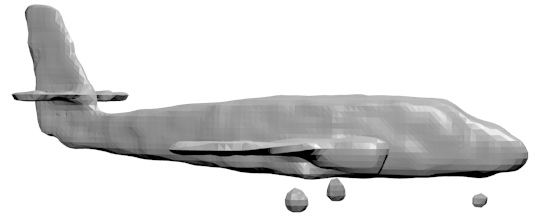} &
    \includegraphics[width=0.2\textwidth]{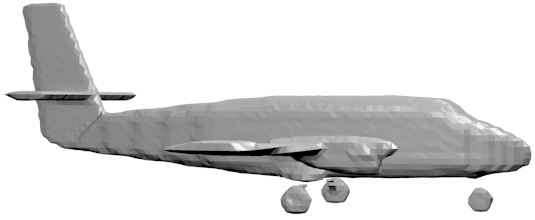} & \\[1.4cm]
    \rotatebox{90}{\small{$256^3$}} & 
    \includegraphics[width=0.2\textwidth]{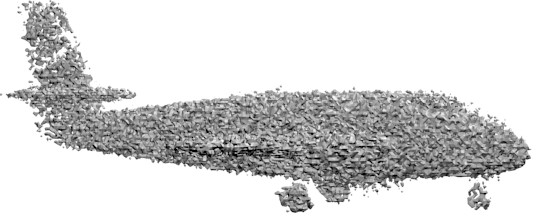} &
    \includegraphics[width=0.2\textwidth]{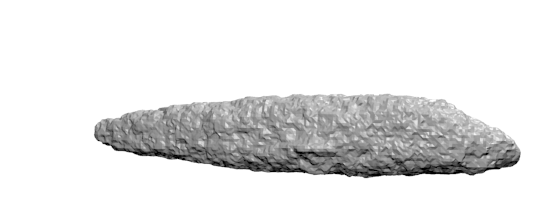} &
    \includegraphics[width=0.2\textwidth]{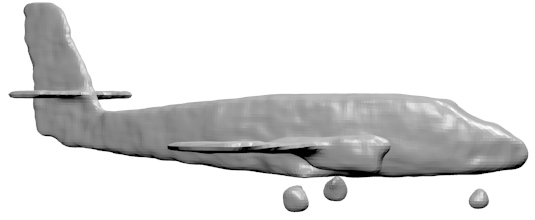} &
    \includegraphics[width=0.2\textwidth]{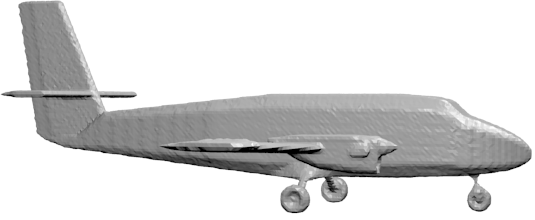} & \\[1.4cm]
    & \small{VolFus \cite{Curless1996SIGGRAPH}}  & \small{TV-L1 \cite{Zach2007ICCV}} & \small{Ours} & \small{Ground Truth} & \\
  \end{tabular}
  \caption{Airplane}
\end{subfigure}
\qquad
\begin{subfigure}{0.5\linewidth}
	\setlength{\tabcolsep}{2pt}
  \begin{tabular}{@{} C{0.05\linewidth} C{0.2\linewidth} C{0.2\linewidth} C{0.2\linewidth} C{0.2\linewidth}}
    \rotatebox{90}{\small{$64^3$}} & 
    \includegraphics[width=0.2\textwidth]{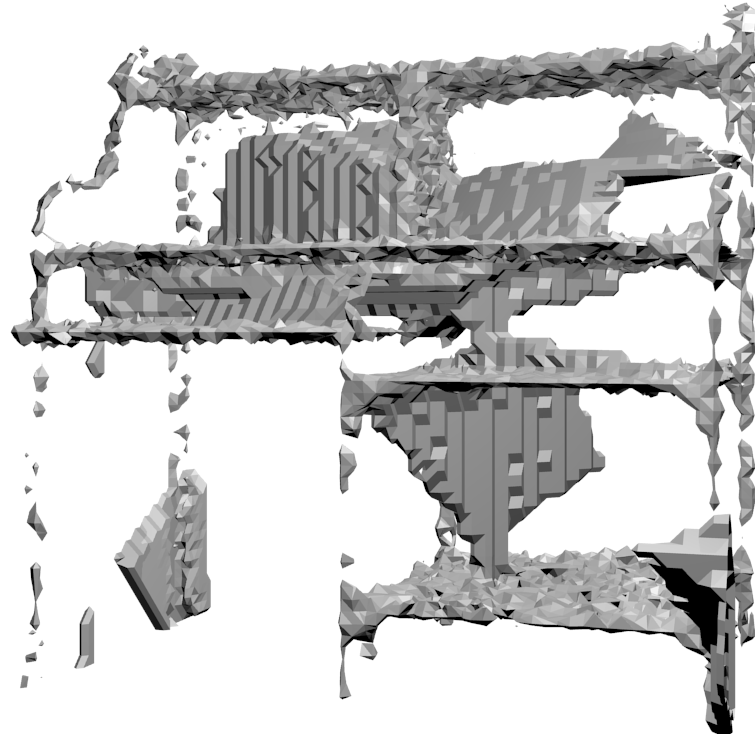} &
    \includegraphics[width=0.2\textwidth]{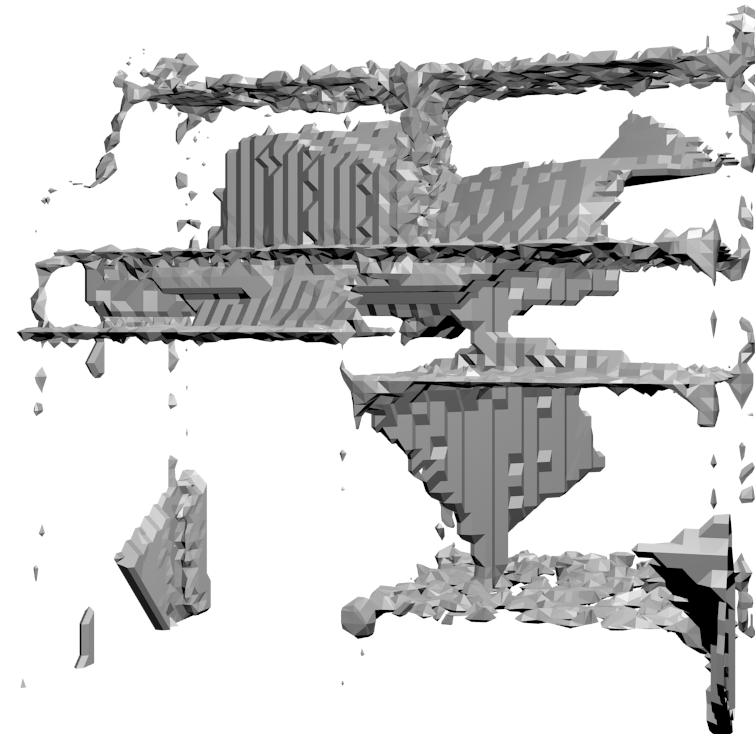} &
    \includegraphics[width=0.2\textwidth]{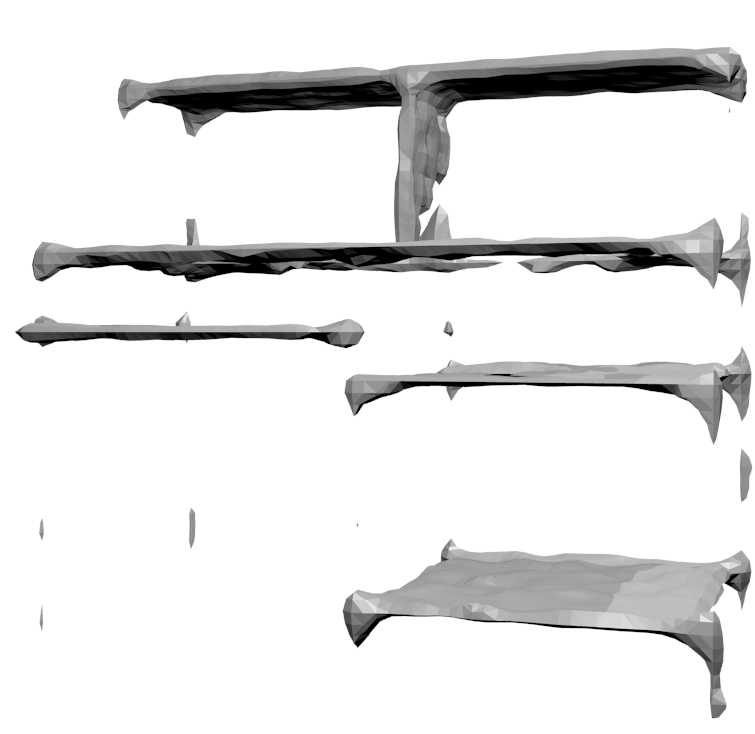} &
    \includegraphics[width=0.2\textwidth]{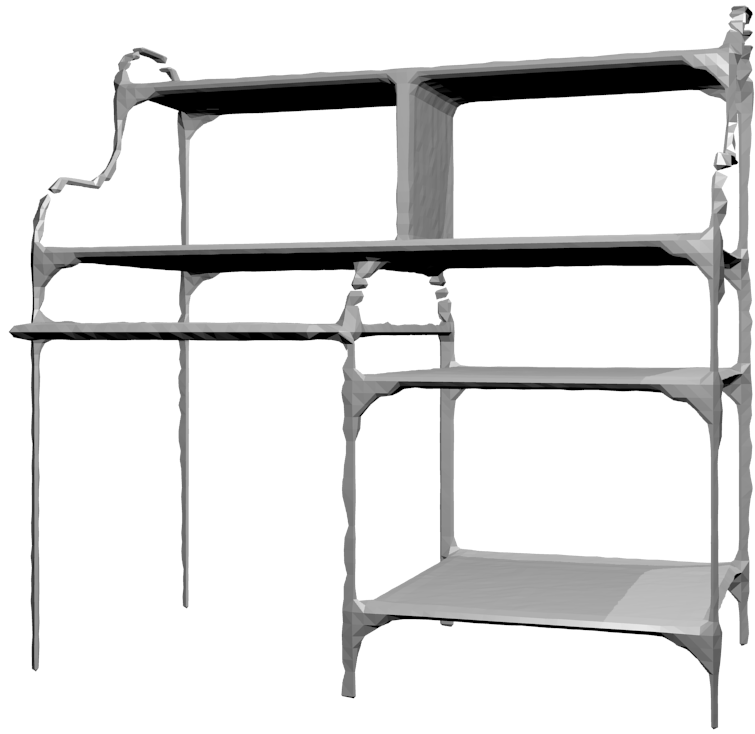} \\
    \rotatebox{90}{\small{$128^3$}} & 
    \includegraphics[width=0.2\textwidth]{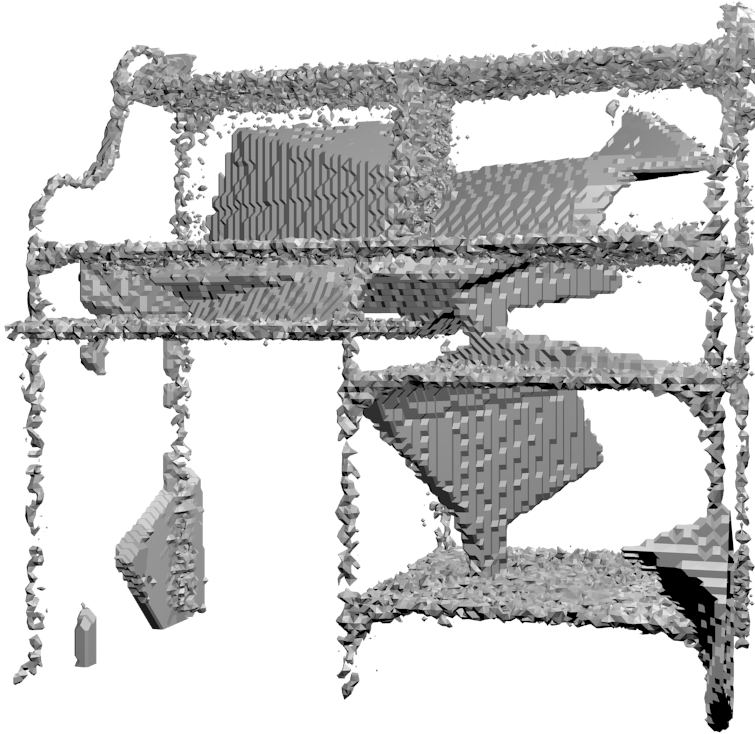} &
    \includegraphics[width=0.2\textwidth]{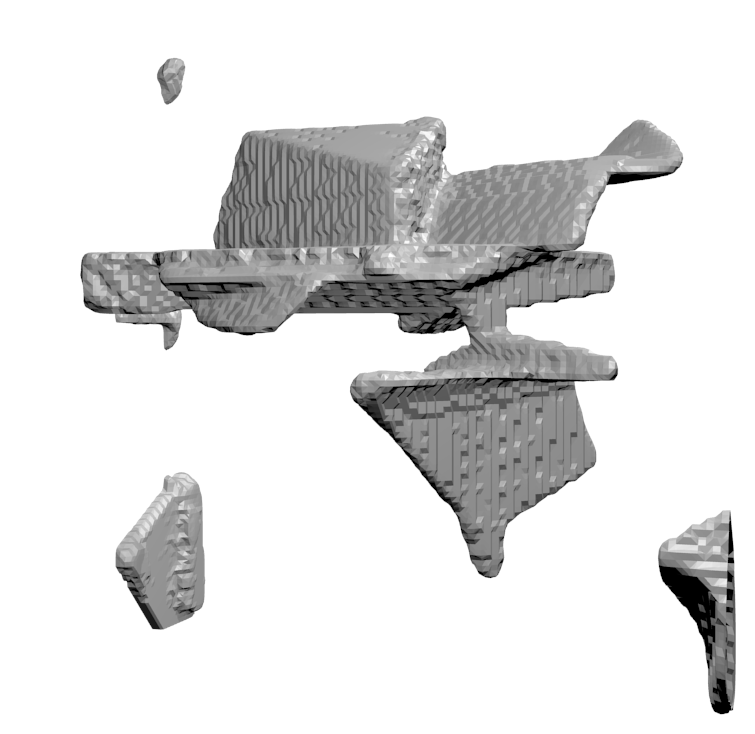} &
    \includegraphics[width=0.2\textwidth]{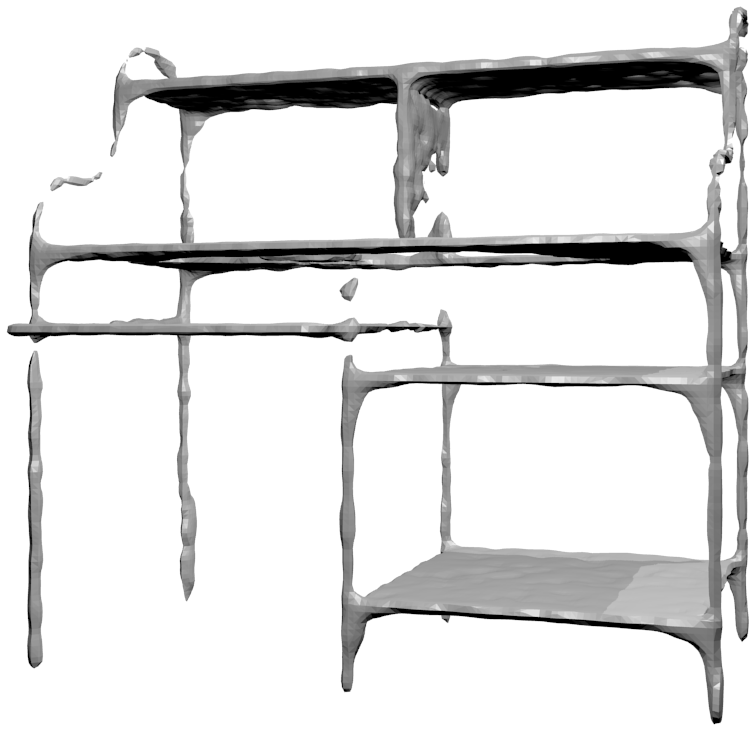} &
    \includegraphics[width=0.2\textwidth]{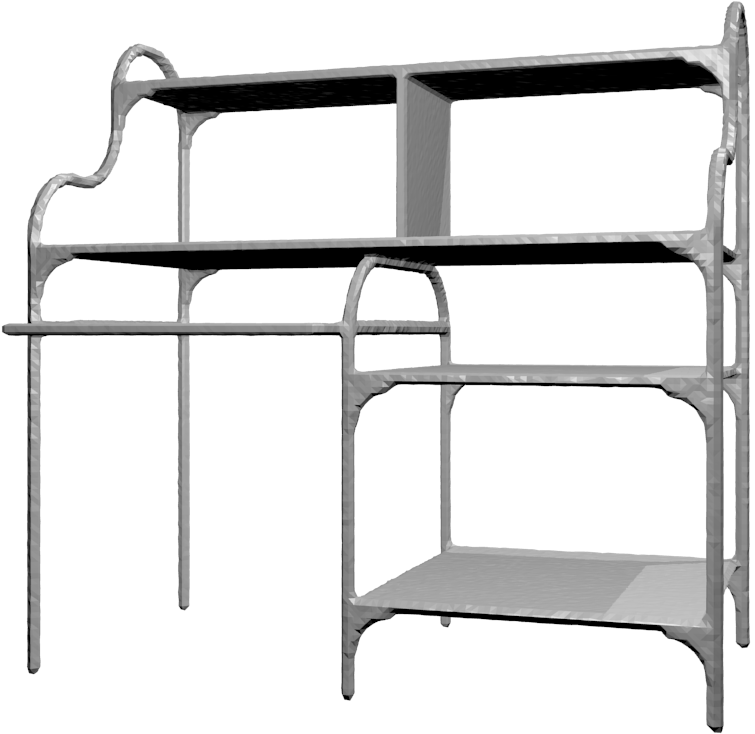} \\
    \rotatebox{90}{\small{$256^3$}} & 
    \includegraphics[width=0.2\textwidth]{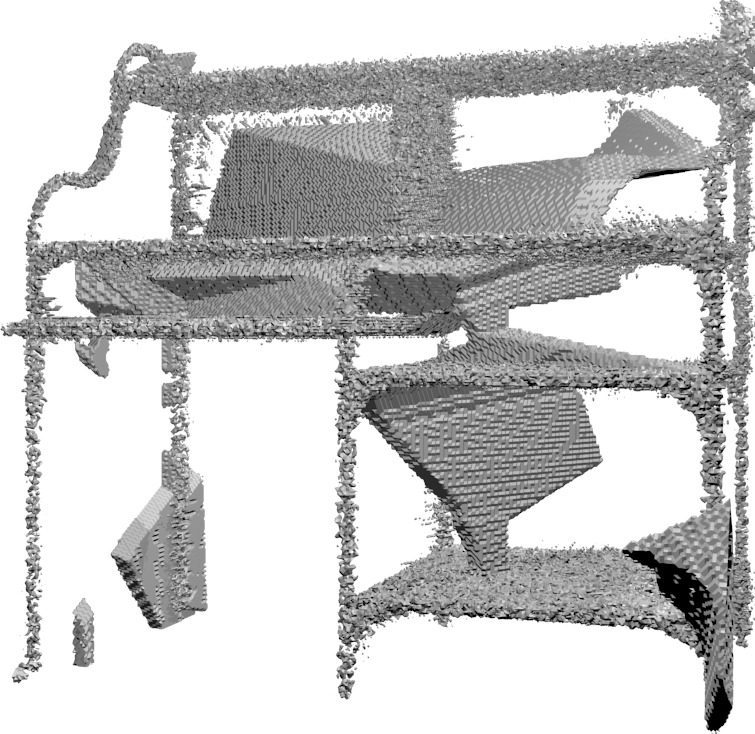} &
    \includegraphics[width=0.2\textwidth]{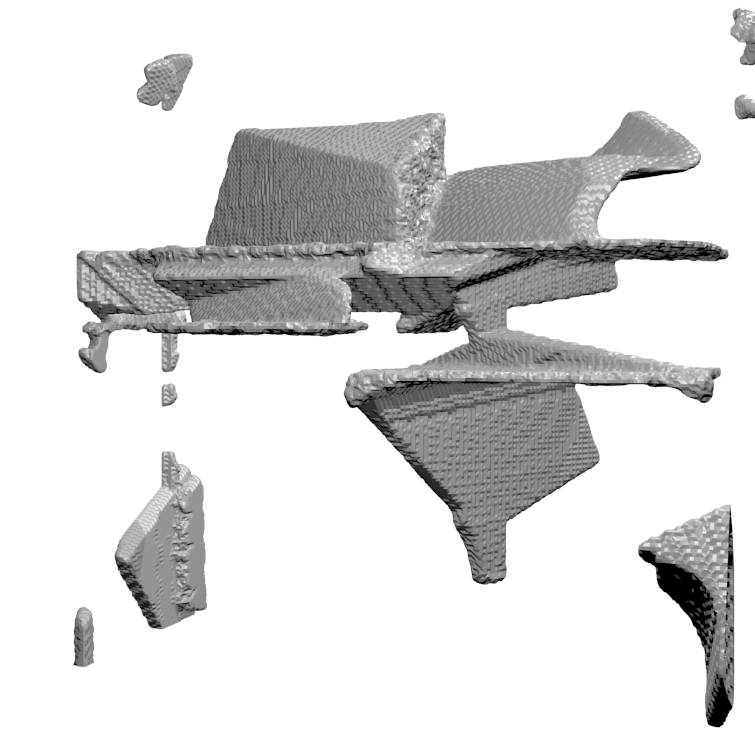} &
    \includegraphics[width=0.2\textwidth]{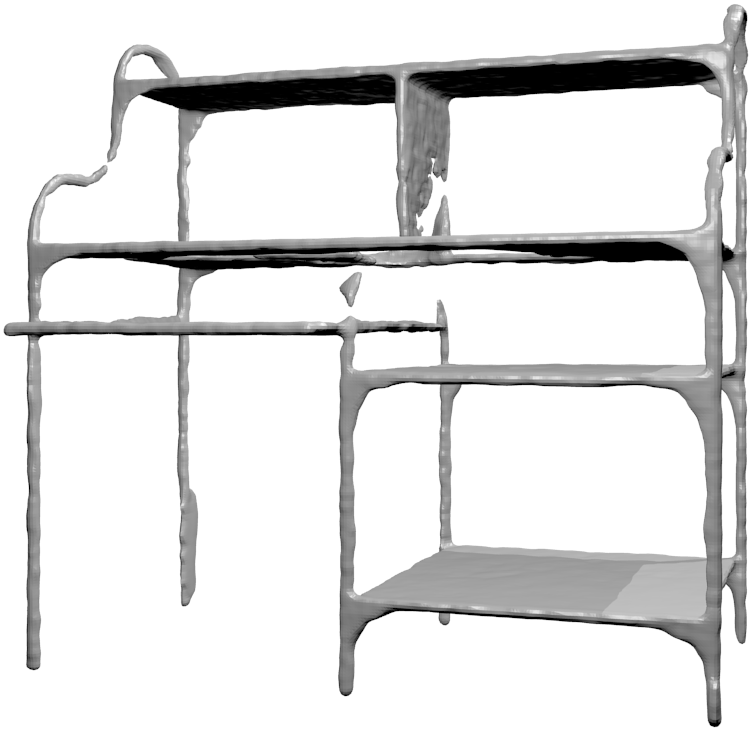} &
    \includegraphics[width=0.2\textwidth]{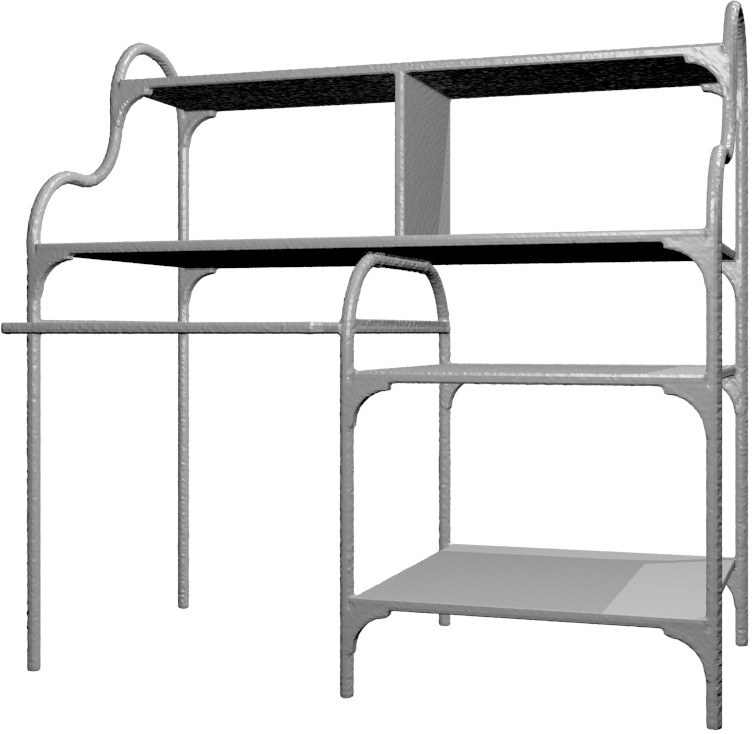} \\
    & \small{VolFus \cite{Curless1996SIGGRAPH}}  & \small{TV-L1 \cite{Zach2007ICCV}} & \small{Ours} & \small{Ground Truth} \\
  \end{tabular}
  \caption{Desk}
\end{subfigure}
\caption{{\bf Qualitative Results on ModelNet} using 4 equally spaced input viewpoints}
\label{fig:mn_qual}
\end{figure*}

\boldparagraph{Qualitative Results on ModelNet}
We show qualitative results on ModelNet in \figref{fig:mn_qual} using the TSDF encoding and $4$ views. The same TSDF truncation threshold has been used for traditional fusion, our OctNetFusion approach and the ground truth generation process. While the baseline approach is not able to resolve conflicting TSDF information from different viewpoints, our approach learns to produce a smooth and accurate 3D model from highly noisy input.

\boldparagraph{Kinect Object Scans}
To evaluate the performance of our algorithm on real data, we use the dataset of Kinect object scans published by Choi \etal \cite{Choi2016ARXIV}. 
For this experiment we vary the number of input views from ten to twenty as the scenes are larger and the camera trajectories are less regular than in the synthetic ModelNet experiments. 
Our results are shown in \tabref{tab:kf_views}. 
While overall errors are larger compared to the ModelNet experiments, we again outperform the traditional fusion approach at all resolutions and for all number of input views. 
Notably, the relative difference between the two methods increases with finer resolution which demonstrates the impact of learned representations for reconstructing details.
In contrast to the experiments on ModelNet, the TV-L1 baseline \cite{Zach2007ICCV} reduces errors more substantially when compared to vanilla volumetric fusion \cite{Curless1996SIGGRAPH}, yet our method is consistently more accurate.

\begin{table}
  \center
    \setlength{\tabcolsep}{3px}
    {\footnotesize \input{tab/kf/sad.tex} }
  \caption{{\bf Fusion of Kinect Object Scans} (MAD in mm)}
  \vspace{-1.5em}
  \label{tab:kf_views}
\end{table}

\boldparagraph{Runtime}
Given its simplicity, Vanilla TSDF fusion~\cite{Curless1996SIGGRAPH} is the fastest method, using just a few milliseconds on a GPGPU.
In contrast, TV-L1 fusion~\cite{Zach2007ICCV} is computationally more expensive. 
We ran all experiments using $700$ iterations of the optimization algorithm.
For an output resolution of $64^3$ TV-L1 needs $0.58$ seconds on average and for an output resolution of $256^3$ it needs $24.66$ seconds on average.
In comparison, our proposed OctNetFusion CNN requires $0.005$ seconds on average for an output resolution of $64^3$ and $10.1$ seconds on average for an output resolution of $256^3$.
All numbers were obtained on a NVidia K80 GPGPU.

\subsection{Volumetric Completion}
\label{sec:volumetric_completion}

\begin{figure}
	\center
	\begin{subfigure}{0.3\linewidth}
		\center
		\includegraphics[width=\linewidth]{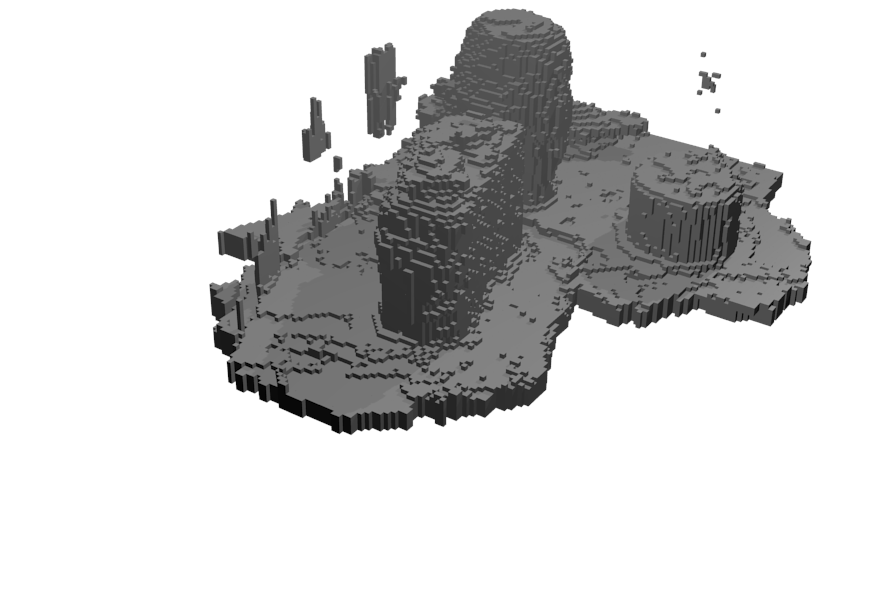} 
		\includegraphics[width=\linewidth]{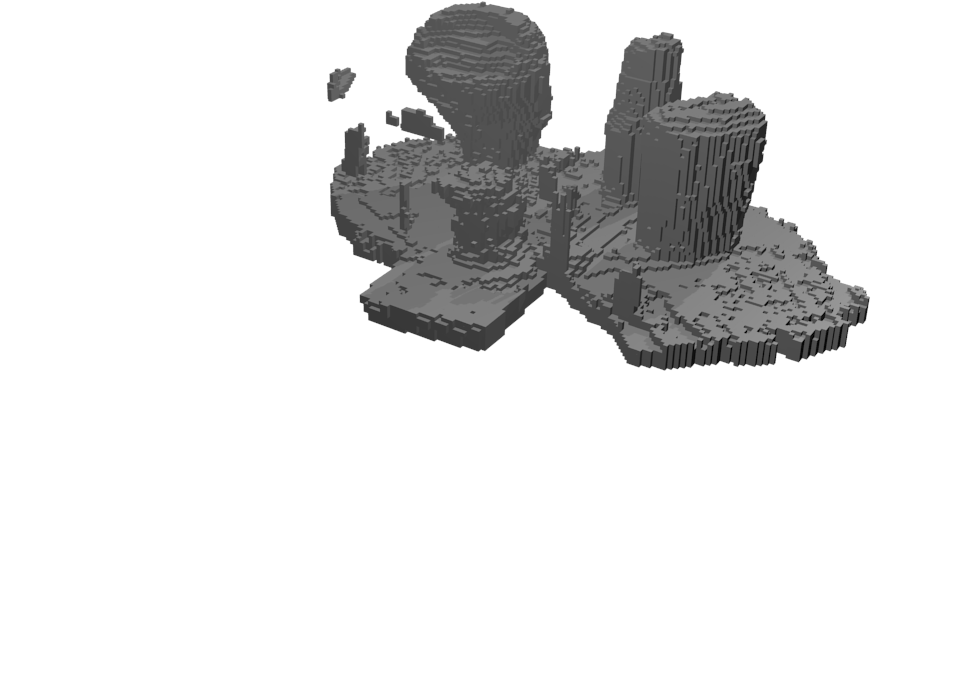} 
		\caption{Firman \cite{Firman2016CVPR}}
	\end{subfigure}
	\begin{subfigure}{0.3\linewidth}
		\center
		\includegraphics[width=\linewidth]{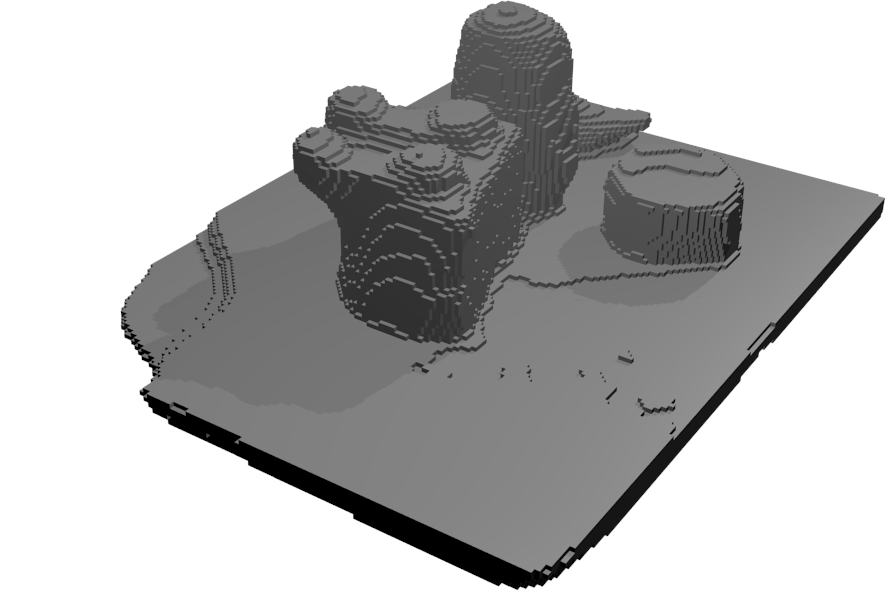} 
		\includegraphics[width=\linewidth]{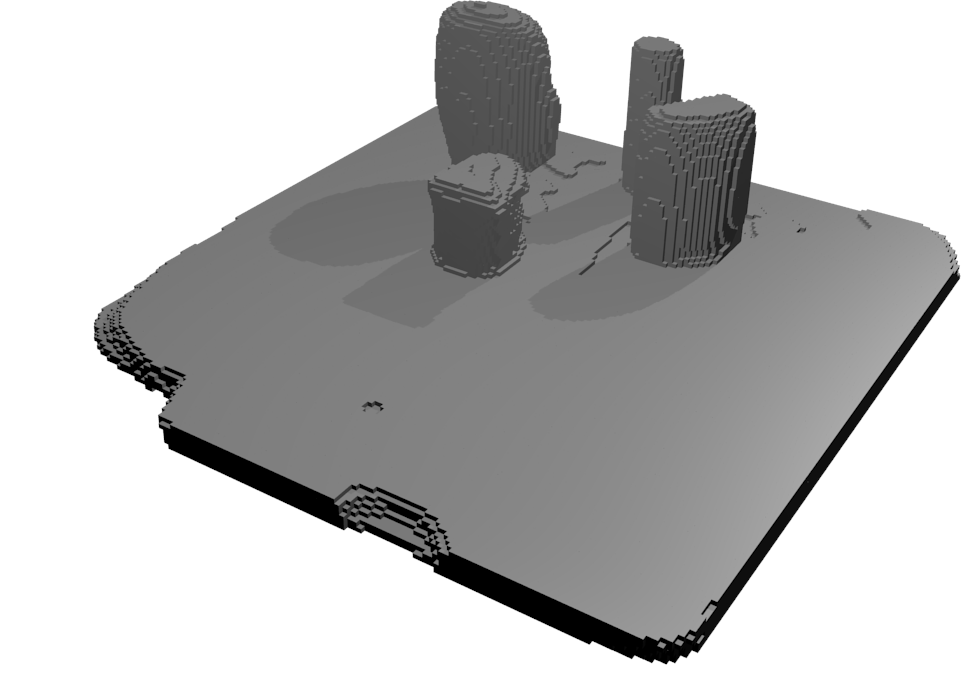} 
		\caption{Ours}
	\end{subfigure} 
	\begin{subfigure}{0.3\linewidth}
		\center
		\includegraphics[width=\linewidth]{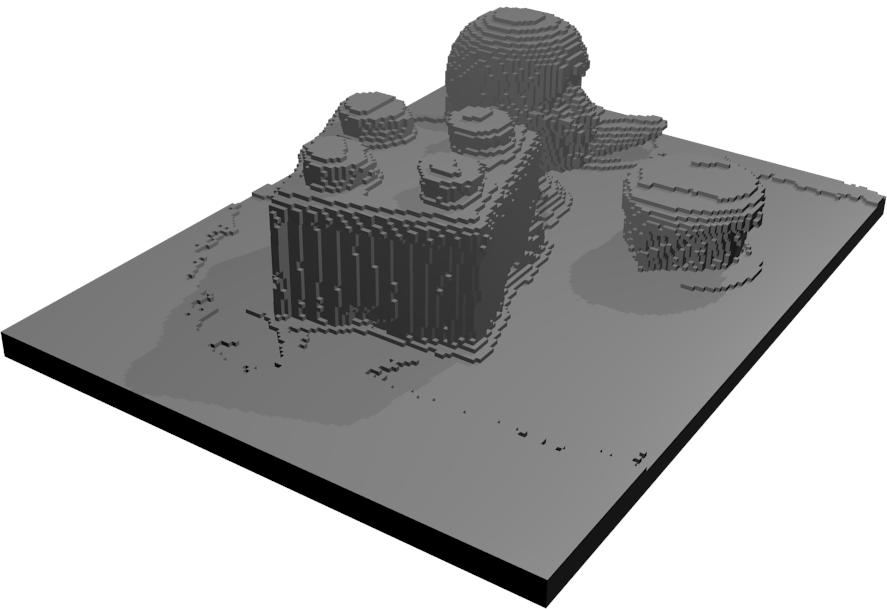} 
		\includegraphics[width=\linewidth]{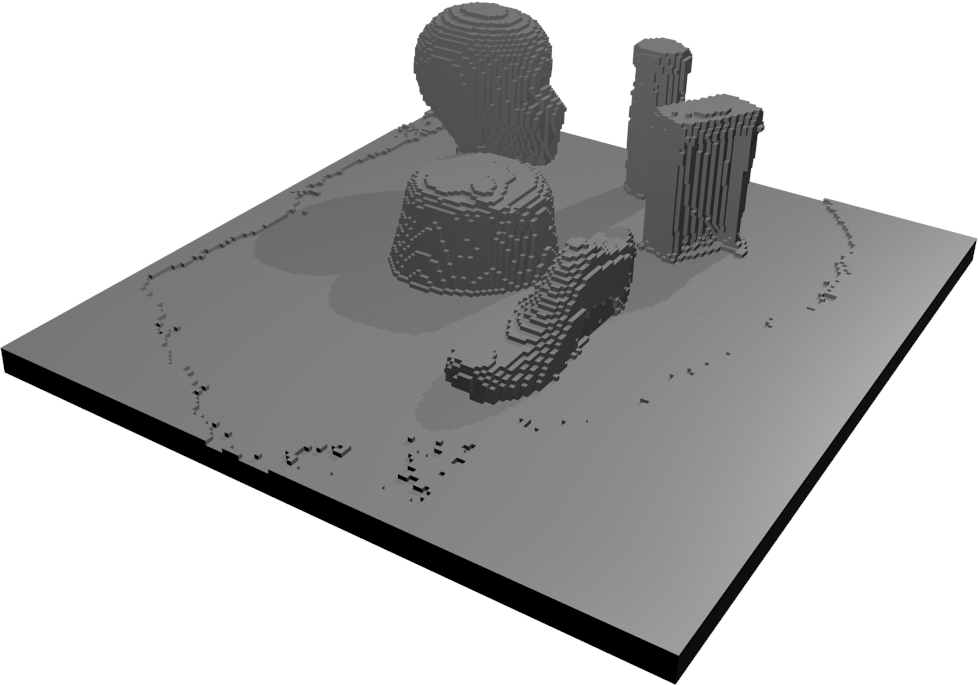} 
		\caption{Ground Truth}
	\end{subfigure}
	\caption{{\bf Volumetric Completion Results}}
	\label{fig:completion}
\end{figure}

In this section, we provide a comparison to Firman's Voxlets approach \etal \cite{Firman2016CVPR} on the task of volumetric shape completion from a single image. 
For this experiment, we use the dataset and metrics proposed by \cite{Firman2016CVPR} and modify our model to predict binary occupancy maps instead of real-valued TSDFs.
Our results are shown in \tabref{tab:firman_tabletop}.
Qualitative results for three different scenes are visualized in \figref{fig:completion}.
As evidenced by our results, our approach improves upon Voxlets \cite{Firman2016CVPR} as well as the method of Zheng \etal \cite{Zheng2013CVPR} in terms of intersection-over-union (IoU) of the occupied space, precision and recall.
Unfortunately, even after communication with the authors we were not able to reproduce their results due to post-publication changes in their dataset.

\begin{table}
\begin{center}
  {\footnotesize \input{tab/voxlets/iou.tex}}
\end{center}
\vspace{-0.8em}
\caption{{\bf Volumetric Completion on Tabletop Dataset.} The numbers marked with asterisk (*) are taken from \cite{Firman2016CVPR} while the others have been recomputed by us and verified by the authors of \cite{Firman2016CVPR}. Unfortunately, even in a joint effort with the authors, we were not able to reproduce the original results due to irreversible post-publication changes in their dataset, thus we provide both results in this table.}
\label{tab:firman_tabletop}
\vspace{-0.8em}
\end{table}

%% file: tab/encodings/sad.tex
\begin{tabular}{lcccccc}
\toprule
 & VolFus \cite{Curless1996SIGGRAPH}  & TV-L1 \cite{Zach2007ICCV} & Occ & TDF + Occ & TSDF & TSDF Hist \\ \midrule
$64^3$ & ${4.136}$ & ${3.899}$ & ${2.095}$ & ${1.987}$ & $\textbf{1.710}$ & ${1.715}$ \\
$128^3$ & ${2.058}$ & ${1.690}$ & ${0.955}$ & ${0.961}$ & ${0.838}$ & $\textbf{0.736}$ \\
$256^3$ & ${1.020}$ & ${0.778}$ & ${0.410}$ & ${0.408}$ & ${0.383}$ & $\textbf{0.337}$ \\
\bottomrule
\end{tabular}

%% file: tab/views/sad.tex
\begin{tabular}{lccc}
\toprule
 & VolFus \cite{Curless1996SIGGRAPH}  & TV-L1 \cite{Zach2007ICCV} & Ours \\ \midrule
1 view & ${14.919}$ & ${14.529}$ & $\textbf{1.927}$ \\
2 views & ${3.929}$ & ${3.537}$ & $\textbf{0.616}$ \\
4 views & ${1.020}$ & ${0.778}$ & $\textbf{0.337}$ \\
6 views & ${0.842}$ & ${0.644}$ & $\textbf{0.360}$ \\
\bottomrule
\end{tabular}

%% file: tab/noise/sad.tex
\begin{tabular}{lcccccccccccc}
\toprule
 & \multicolumn{3}{c}{$\sigma=0.00$} & \multicolumn{3}{c}{$\sigma=0.01$} & \multicolumn{3}{c}{$\sigma=0.02$} & \multicolumn{3}{c}{$\sigma=0.03$} \\ \cmidrule(lr){2-4} \cmidrule(lr){5-7} \cmidrule(lr){8-10} \cmidrule(lr){11-13}
 & VolFus \cite{Curless1996SIGGRAPH}  & TV-L1 \cite{Zach2007ICCV} & Ours & VolFus \cite{Curless1996SIGGRAPH}  & TV-L1 \cite{Zach2007ICCV} & Ours & VolFus \cite{Curless1996SIGGRAPH}  & TV-L1 \cite{Zach2007ICCV} & Ours & VolFus \cite{Curless1996SIGGRAPH}  & TV-L1 \cite{Zach2007ICCV} & Ours \\ \midrule
$64^3$  & ${3.020}$ & ${3.272}$ & $\textbf{1.647}$ & ${3.439}$ & ${3.454}$ & $\textbf{1.487}$ & ${4.136}$ & ${3.899}$ & $\textbf{1.715}$ & ${4.852}$ & ${4.413}$ & $\textbf{1.938}$ \\
$128^3$ & ${1.330}$ & ${1.396}$ & $\textbf{0.744}$ & ${1.647}$ & ${1.543}$ & $\textbf{0.676}$ & ${2.058}$ & ${1.690}$ & $\textbf{0.736}$ & ${2.420}$ & ${1.850}$ & $\textbf{0.804}$ \\
$256^3$ & ${0.621}$ & ${0.637}$ & $\textbf{0.319}$ & ${0.819}$ & ${0.697}$ & $\textbf{0.321}$ & ${1.020}$ & ${0.778}$ & $\textbf{0.337}$ & ${1.188}$ & ${0.858}$ & $\textbf{0.402}$ \\
\bottomrule
\end{tabular}

%% file: tab/classgeneralization/sad.tex
\begin{tabular}{llcccc}
\toprule
 & & VolFus \cite{Curless1996SIGGRAPH} & TV-L1 \cite{Zach2007ICCV} & Seen & Unseen \\ \midrule
\multirow{3}{*}{all}       & $64^3$  & ${4.136}$ & ${3.899}$ & ${1.715}$ & $\textbf{1.686}$ \\
                           & $128^3$ & ${2.058}$ & ${1.690}$ & $\textbf{0.736}$ & ${0.799}$ \\
                           & $256^3$ & ${1.020}$ & ${0.778}$ & $\textbf{0.337}$ & ${0.358}$ \\ \midrule
\multirow{3}{*}{airplane}  & $64^3$  & ${0.668}$ & ${0.583}$ & $\textbf{0.419}$ & ${0.470}$ \\
                           & $128^3$ & ${0.324}$ & ${0.297}$ & $\textbf{0.174}$ & ${0.192}$ \\
                           & $256^3$ & ${0.157}$ & ${0.111}$ & $\textbf{0.076}$ & $\textbf{0.076}$ \\ \midrule
\multirow{3}{*}{desk}      & $64^3$  & ${5.122}$ & ${4.767}$ & $\textbf{1.954}$ & ${2.000}$ \\
                           & $128^3$ & ${2.540}$ & ${2.165}$ & $\textbf{0.777}$ & ${0.898}$ \\
                           & $256^3$ & ${1.260}$ & ${0.987}$ & $\textbf{0.334}$ & ${0.383}$ \\
\bottomrule
\end{tabular}

%% file: tab/kf/sad.tex
\begin{tabular}{lcccccc}
\toprule
 & \multicolumn{3}{c}{views=10} & \multicolumn{3}{c}{views=20} \\ \cmidrule(lr){2-4} \cmidrule(lr){5-7}
 & VolFus \cite{Curless1996SIGGRAPH}  & TV-L1 \cite{Zach2007ICCV} & Ours & VolFus \cite{Curless1996SIGGRAPH}  & TV-L1 \cite{Zach2007ICCV} & Ours \\ \midrule
$64^3$  & ${103.855}$ & ${25.976}$ & $\textbf{22.540}$  & ${72.631}$ & ${22.081}$ & $\textbf{18.422}$ \\
$128^3$ & ${58.802}$  & ${12.839}$ & $\textbf{11.827}$  & ${41.631}$ & ${11.924}$ & $\textbf{9.637}$ \\
$256^3$ & ${31.707}$  & ${5.372}$  & $\textbf{4.806}$   & ${22.555}$ & ${5.195}$  & $\textbf{4.110}$ \\
\bottomrule
\end{tabular}

%% file: tab/voxlets/iou.tex
\begin{tabular}{lccc}
\toprule
Method & IoU & Precision & Recall\\
\midrule\vspace{-1em}\\
Zheng \etal \cite{Zheng2013CVPR}*  & 0.528 & 0.773 & 0.630\\
Voxlets \etal \cite{Firman2016CVPR}*  & 0.585 & 0.793 & 0.658\\
\midrule\vspace{-1em}\\
Voxlets \etal \cite{Firman2016CVPR}  & 0.550 & 0.734 & 0.705\\
Ours  & {\bf 0.650} & {\bf 0.834} & {\bf 0.756}\\
\bottomrule
\end{tabular}

%% file: sec_conclusion.tex
\section{Conclusion} 
\label{sec:conclusion}

We propose OctNetFusion, a deep 3D convolutional neural network that is capable of fusing depth information from different viewpoints to produce accurate and complete 3D reconstructions. Our experiments demonstrate the advantages of our learning-based approach over the traditional fusion baseline and show that our method generalizes to novel object categories, producing compelling results on both synthetic and real Kinect data. While in this paper we have focused on the problem of fusing depth maps, an interesting direction for future work is to extend our model to 3D reconstruction from RGB images where 3D representations are learned jointly with 2D image representations in an end-to-end fashion.